\documentclass[letterpaper, 10 pt, conference]{ieeeconf}
\IEEEoverridecommandlockouts
\usepackage[colorlinks,bookmarksopen,bookmarksnumbered,citecolor=red,urlcolor=red]{hyperref}
\usepackage[pdftex]{graphicx}
\usepackage{color}
\usepackage[binary-units=true]{siunitx}
\usepackage{amsfonts}
\usepackage{glossaries}
\glsdisablehyper
\hypersetup
{
	pdftitle = {An efficient and versatile framework for multi-contact optimal control},
	pdfauthor = {Carlos Mastalli},
	pdfsubject = {Submitted to Robotics and Automation Letters (RAL)},
  pdfkeywords = {optimal control, differential dynamic programming, multiple-shooting, legged robotics},
	pdftoolbar = true,
	colorlinks = true,
	linkcolor = black,
	citecolor = black,
	urlcolor = black,
}
\newacronym{fonc}{FONC}{First-order Necessary Condition}
\newacronym{oc}{OC}{Optimal Control}
\newacronym{mpc}{MPC}{Model Predictive Control}
\newacronym{mcoc}{MCOC}{Multi-Contact Optimal Control}
\newacronym{lqr}{LQR}{Linear-Quadratic Regulator}
\newacronym{qp}{QP}{Quadratic Programming}
\newacronym{lq}{LQ}{Linear Quadratic}
\newacronym{slq}{SLQ}{Sequential Linear Quadratic}
\newacronym{sqp}{SQP}{Sequential Quadratic Programming}
\newacronym{ddp}{DDP}{Differential Dynamic Programming}
\newacronym{fddp}{FDDP}{Feasibility-driven Differential Dynamic Programming}
\newacronym{com}{CoM}{Center of Mass}
\newacronym{crocoddyl}{Crocoddyl}{Contact RObot COntrol by Differential DYnamic Library}
\newacronym{am}{AM}{Action Model}
\newacronym{dam}{DAM}{Differential Action Model}
\newacronym{iam}{IAM}{Integrated Action Model}
\newacronym{kkt}{KKT}{Karush-Kuhn-Tucker}
\newacronym{rnea}{RNEA}{Recursive Newton-Euler Algorithm}
\newacronym{aba}{ABA}{Articulated Body Algorithm}
\newacronym{simd}{SIMD}{Same Instruction Multiple Data}
\newcommand{\sref}[1]{Section~\ref{#1}}
\newcommand{\fref}[1]{Fig.~\ref{#1}}
\newcommand{\eref}[1]{Eq.~(\ref{#1})}

\title{\LARGE \bf Crocoddyl: An Efficient and Versatile Framework\\ for Multi-Contact Optimal Control}

\author{
Carlos Mastalli\,$^{1,2,3}$\quad Rohan Budhiraja\,$^1$\quad Wolfgang Merkt\,$^{2,4}$\quad Guilhem Saurel\,$^1$\quad Bilal Hammoud\,$^{5,6}$\quad\\
Maximilien Naveau\,$^5$\quad Justin Carpentier\,$^7$\quad Ludovic Righetti\,$^{5,6}$\quad Sethu Vijayakumar\,$^{2,3}$\quad Nicolas Mansard\,$^1$%
\thanks{$^1$~Gepetto Team, LAAS-CNRS, Toulouse, France.}
\thanks{$^2$~School of Informatics, University of Edinburgh, Edinburgh, UK.}
\thanks{$^3$~The Alan Turing Institute, Edinburgh, UK.}
\thanks{$^4$~Oxford Robotics Institute, University of Oxford, UK.}
\thanks{$^5$~Max Planck Institute for Intelligent Systems, T\"ubingen, Germany.}
\thanks{$^6$~Tandon School of Engineering, New York University, USA.}
\thanks{$^7$~INRIA, ENS, CNRS, PSL Research University, Paris, France.}
\thanks{\textit{email}: \href{mailto:carlos.mastalli@ed.ac.uk}{carlos.mastalli@ed.ac.uk.}
This research was supported by (1) the European Commission under the Horizon 2020 project Memory of Motion (MEMMO, project ID: 780684), (2) the Engineering and Physical Sciences Research Council (EPSRC) UK RAI Hub for Offshore Robotics for Certification of Assets (ORCA, grant reference EP/R026173/1), (3) the European Research Council grant No 63793, (4) the US National Science Foundation under grant CMMI-1825993 and (5) the European Flag-ERA JTC Project RobCom++.}
}

\begin{document}
\maketitle

\begin{abstract}
  We introduce \acrshort{crocoddyl} (\acrlong{crocoddyl}), an open-source framework tailored for efficient multi-contact optimal control.
  \acrshort{crocoddyl} efficiently computes the state trajectory and the control policy for a given predefined sequence of contacts.
  Its efficiency is due to the use of sparse analytical derivatives, exploitation of the problem structure, and data sharing.
  It employs differential geometry to properly describe the state of any geometrical system, e.g. floating-base systems.
  Additionally, we propose a novel optimal control algorithm called \gls{fddp}.
  Our method does not add extra decision variables which often increases the computation time per iteration due to factorization.
  \gls{fddp} shows a greater globalization strategy compared to classical \gls{ddp} algorithms.
  Concretely, we propose two modifications to the classical \gls{ddp} algorithm.
  First, the backward pass accepts infeasible state-control trajectories.
  Second, the rollout keeps the gaps open during the early ``exploratory'' iterations (as expected in multiple-shooting methods with only equality constraints).
  We showcase the performance of our framework using different tasks.
  With our method, we can compute highly-dynamic maneuvers (e.g. jumping, front-flip) within few milliseconds.
\end{abstract}

\section{Introduction}\label{sec:introduction}
Multi-contact optimal control promises to generate whole-body motions and control policies that allow legged robots to robustly react to unexpected events in real-time.
It has several advantages compared with state-of-the-art frameworks (e.g.~\cite{bellicoso-iros-17,mastalli-17unpublished}) in which a whole-body controller (e.g.~\cite{herzog-iros14,focchi-ar17,mastalli-ral18}) compliantly tracks an optimized Centroidal dynamics trajectory (e.g.~\cite{ponton-ichr16,carpentier-tro18}) with optionally an optimized contact plan (e.g.~\cite{aceituno_cabezas-ral17,winkler-ral18,mastalli-icra17}).
For instance, they cannot properly handle the robot orientation, particularly during flight phases due to the nonholonomic effect on the dynamics, and to regulate the angular momentum to zero leads to tracking errors even in walking motions~\cite{wieber-fmbr05}.
Furthermore, it is well-known that instantaneous time-invariant control (i.e. \textit{instantaneous} whole-body control) cannot properly track nonholonomic systems~\cite{brockett-dgct83}.
Indeed, in our previous work~\cite{budhiraja-ichr18}, we have shown that whole-body planning produces more efficient motions, with lower forces and impacts.

\begin{figure}
    \centering
    \href{https://youtu.be/wHy8YAHwj-M?t=56}{\includegraphics[width=0.87\linewidth]{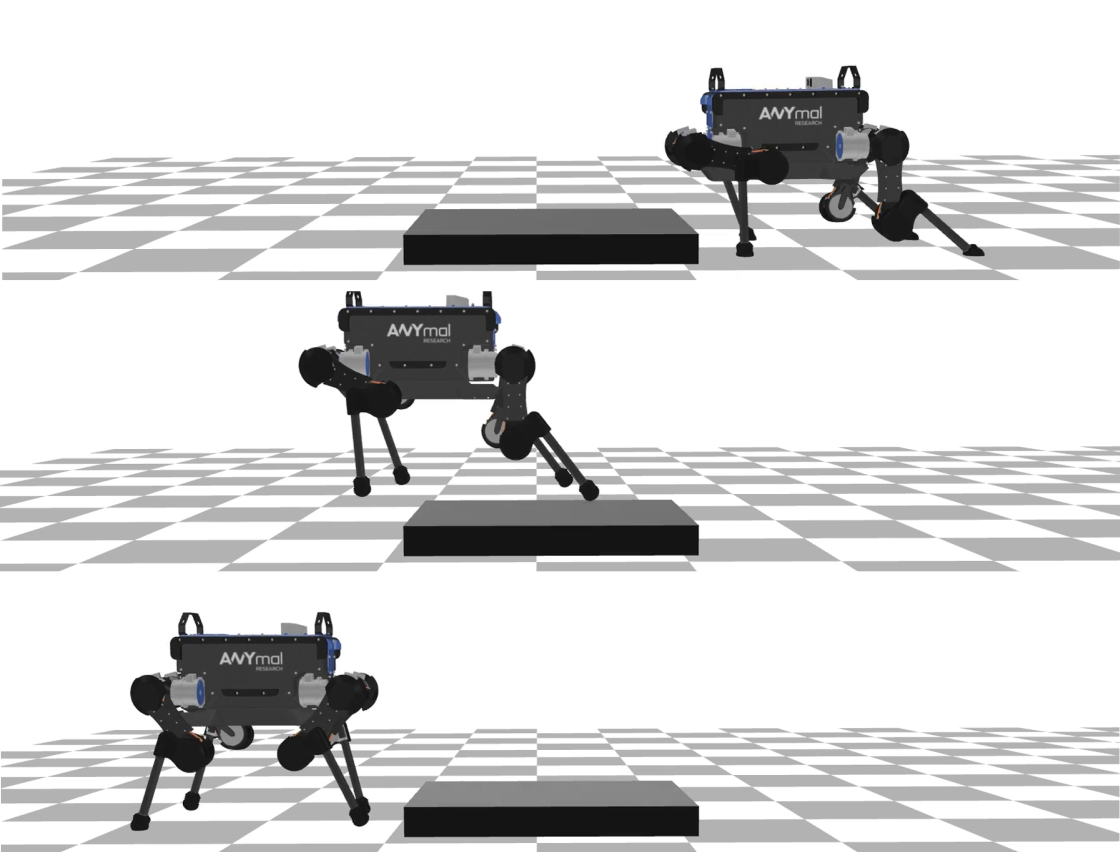}}
    \caption{Crocoddyl: an efficient and versatile framework for multi-contact optimal control~\cite{crocoddylweb}.
    Highly-dynamic maneuvers are needed to traverse an obstacle with the ANYmal robot.}
    \label{fig:cover_fig}
\end{figure}

Recent work on optimal control has shown that nonlinear \gls{mpc} is plausible for controlling legged robots in real-time \cite{koenemann-iros15,neunert-ral18,dicarlo-iros18}.
All these methods have in common that they solve the nonlinear \gls{oc} problem by iteratively building and solving a \gls{lqr} problem (i.e. \gls{ddp} with Gauss-Newton approximation~\cite{li-icinco04}).
These frameworks use numerical or automatic differentiation which is often inefficient compared to sparse and analytical derivatives~\cite{carpentier-rss18}.
Furthermore, they do not explicitly handle the geometric structure of legged systems which include elements of $\mathbb{SE}(3)$.
\gls{ddp} has proven to efficiently solve nonlinear \gls{oc} problems due to its intrinsic sparse structure.
However, it has poor globalization strategy and struggles to handle infeasible warm-start\footnote{An infeasible warm-start refers to state and control trajectories that are not consistent with the system dynamics.}.
In this vein, Giftthaler et al.~\cite{giftthaler-iros18} proposed a variant of the \gls{ddp} algorithm for multiple-shooting \gls{oc}, which has a better convergence rate than \gls{ddp}.
Nonetheless, the gap contraction rate does not numerically match the \gls{kkt} problem applied to the multiple-shooting formulation with only equality constraints~\cite{bock-ifac84}.
In this work, we address these drawbacks by computing highly-dynamic maneuvers as shown in~\fref{fig:cover_fig}.

\subsection{Contribution}
We propose a novel and efficient framework for multi-contact \gls{oc} called \acrshort{crocoddyl}.
Our framework efficiently solves this problem by employing sparse and analytical derivatives of the contact and impulse dynamics.
The \gls{oc} solver properly handles the geometry of rigid bodies using dedicated numerical routines for Lie groups and their derivatives.
Indeed, we model the floating-base as a $\mathbb{SE}(3)$ element, needed for example for the generation of front-flip motions.
Additionally, we propose a variant of the \gls{ddp} algorithm that matches the behavior of the Newton method applied to the \gls{kkt} conditions of a direct multiple-shooting formulation with only equality constraints.
Our algorithm is called \acrlong{fddp} (\gls{fddp})\footnote{We also refer as feasibility-prone \gls{ddp}.} as it handles infeasible guesses that occur whenever there is a gap between subsequent nodes in the trajectory. 
\gls{fddp} has a greater globalization strategy compared to classical \gls{ddp}, allowing us to solve complex maneuvers in few iterations.
\section{Multi-contact Optimal Control}\label{sec:mcop}
In this section, we first introduce the multi-contact optimal control problem for multibody systems under physical constraints (\sref{sec:multibody_ocp}).
We simplify the problem by modeling contacts as holonomic constraints (\sref{sec:holonomic_ocp}).
With this method, we derive tailored analytical and sparse derivatives for fast computation.
The calculation of derivatives typically represents the main computation carried out by optimal control solvers.

\subsection{Formulation of the optimal control problem}\label{sec:multibody_ocp}
We focus on an efficient formulation of the multi-contact optimal control problem.
One can formulate this problem as follows:
\begin{equation}
    \label{eq:oc_problem}
    \begin{aligned}
        \begin{Bmatrix} \mathbf{x}^*_0,\cdots,\mathbf{x}^*_{N} \\
                        \mathbf{u}^*_0,\cdots,\mathbf{u}^*_{N-1}
        \end{Bmatrix} =
        \arg\min_{\mathbf{X},\mathbf{U}}
        &
        & & \hspace{-0.75em}l_N(\mathbf{x}_{N})+\sum_{k=0}^{N-1} \int_{t_k}^{t_k+\Delta t_k}\hspace{-2.em} l(\mathbf{x},\mathbf{u})dt \\
        & \hspace{-2em}\textrm{s.t.}
        & &   \hspace{-1em}\dot{\mathbf{v}}, \boldsymbol{\lambda} = \arg\min_{\dot{\mathbf{v}},\boldsymbol{\lambda}} \|\dot{\mathbf{v}} - \dot{\mathbf{v}}_{free}\|_{\mathbf{M}},\\
        & & & \mathbf{x}\in\mathcal{X}, \mathbf{u}\in\mathcal{U}
    \end{aligned}
\end{equation}
where the state $\mathbf{x}=(\mathbf{q},\mathbf{v})\in X$ lies on a differential manifold formed by the configuration point $\mathbf{q}$ and its tangent vector $\mathbf{v}$ and is described by a $n_x$-tuple, the control $\mathbf{u}=(\boldsymbol{\tau},\boldsymbol{\lambda})\in\mathbb{R}^{n_u}$ composed by input torque commands $\boldsymbol{\tau}$ and contact forces $\boldsymbol{\lambda}$, $\dot{\mathbf{x}}\in T_\mathbf{x} X$ lies in the tangent space of the state manifold and it is described by a $n_{dx}$-tuple, and $\mathcal{X}$, $\mathcal{U}$ represent the state and control admissible sets, respectively, $\dot{\mathbf{v}}_{free}$ is the unconstrained acceleration in generalized coordinates, and $\mathbf{M}$ is the joint-space inertia matrix.

This problem can be seen as a bilevel optimization, where the lower-level optimization uses the Gauss principle of least constraint to describe the physical constraints as described in~\cite{kalaba-94}.
State and control admissible sets can belong to the lower-level optimization (e.g., joint limits and force friction constraints) as well as to the upper-level one (e.g., task-related constraints and collision with the environment).

\subsection{Contacts as holonomic constraints}\label{sec:holonomic_ocp}
To solve this optimization problem in real-time, we need to efficiently handle (a) the high-dimensionality of the search-space and (b) the instabilities, discontinuities, and non-convexity of the system dynamics (lower-level optimization), among others.
One way of reducing the complexity of the \gls{oc} problem is by solving the lower-level optimization analytically, e.g.~\cite{budhiraja-ichr18}.
Indeed, we have implemented the contact model using holonomic scleronomic constraints on the frame placement (i.e. $\boldsymbol{\phi}(\mathbf{q})=\mathbf{0}$ where $\mathbf{J}_c = \frac{{\partial\boldsymbol{\phi}}}{\partial{\mathbf{q}}}$ is the contact Jacobian) as:
\begin{equation}\label{eq:kkt_fwddyns}
    \left[\begin{matrix}\dot{\mathbf{v}} \\ -\boldsymbol{\lambda}\end{matrix}\right] =
    \left[\begin{matrix}\mathbf{M} & \mathbf{J}^{\top}_c \\ {\mathbf{J}_{c}} & \mathbf{0} \end{matrix}\right]^{-1}
    \left[\begin{matrix}\boldsymbol{\tau}_b \\ -\mathbf{a}_0 \\\end{matrix}\right] =
    \left[\begin{matrix}\mathbf{y}(\mathbf{x},\boldsymbol{\tau}) \\ -\mathbf{g}(\mathbf{x},\boldsymbol{\tau})\end{matrix}\right],
\end{equation}
where $\mathbf{J}_c$ is expressed in the local frame, and $\mathbf{a}_0\in\mathbb{R}^{n_f}$ is the desired acceleration in the constraint space.
\eref{eq:kkt_fwddyns} allows us to express the contact forces in terms of the state and torques, and it has a unique solution if $\mathbf{J}_c$ is full-rank.
To improve stability in the numerical integration, we define PD gains that are similar in spirit to Baumgarte stabilization~\cite{baumgarte-72}:
\begin{equation}\label{eq:baumgarte_stabilization}
    \mathbf{a}_0 = \mathbf{a}_{\lambda(c)} - \alpha \,^oM^{ref}_{\lambda(c)}\ominus\,^oM_{\lambda(c)} - \beta\mathbf{v}_{\lambda(c)},
\end{equation}
where $\mathbf{v}_{\lambda(c)}$, $\mathbf{a}_{\lambda(c)}$ are the spatial velocity and acceleration at the parent body of the contact $\lambda(c)$, respectively, $\alpha$ and $\beta$ are the stabilization gains, and $\,^oM^{ref}_{\lambda(c)}\ominus\,^oM_{\lambda(c)}$ is the $\mathbb{SE}(3)$ inverse composition between the reference contact placement and the current one~\cite{blanco-10se3}.

As \eref{eq:kkt_fwddyns} neglects the friction-cone constraints and the joint limits, the dynamics describe an equality constraint and we can use an unconstrained \gls{ddp} solver~\cite{mayne-66}.
Nonetheless, inequality constraints can still be included in \gls{ddp}-like solvers, i.e. using
penalization, active-set~\cite{xie-icra17}, or Augmented Lagrangian~\cite{howell-19} strategy.

\subsubsection{Efficient rollout and derivative computation}
We do not need to invert the entire \gls{kkt} matrix in \eref{eq:kkt_fwddyns} during the numerical integration of the dynamics.
Indeed, the evolution of the system acceleration and contact can be described as:
\begin{eqnarray}
    \mathbf{y}(\mathbf{x},\boldsymbol{\tau}) &=& \mathbf{M}^{-1}\left(\boldsymbol{\tau}_b + \mathbf{J}^\top_c\mathbf{g}(\mathbf{x},\boldsymbol{\tau})\right),\nonumber\\
    \mathbf{g}(\mathbf{x},\boldsymbol{\tau}) &=& \widehat{\mathbf{M}}^{-1}(\mathbf{a}_0 - \mathbf{J}_c\mathbf{M}^{-1}\boldsymbol{\tau}_b),
\end{eqnarray}
and, for instance, we can use the Cholesky decomposition for efficiently computing $\mathbf{M}^{-1}$ and $\widehat{\mathbf{M}}^{-1}=\mathbf{J}_c\mathbf{M}^{-1}\mathbf{J}^\top_c$.
Note that $\widehat{\mathbf{M}}$ is the operational space inertia matrix~\cite{khatib-jra87}.

If we analytically derive \eref{eq:kkt_fwddyns} by applying the chain rule, then we can describe the Jacobians of $\mathbf{y}(\cdot)$ and $\mathbf{g}(\cdot)$ with respect to the derivatives of the \gls{rnea} algorithm and kinematics, i.e.:
\begin{eqnarray}\label{eq:kkt_fwddyns_der}
\left[\begin{matrix}\delta\dot{\mathbf{v}} \\ -\delta\boldsymbol{\lambda}\end{matrix}\right] &=&
-\left[\begin{matrix}\mathbf{M} & \mathbf{J}^{\top}_c \\ {\mathbf{J}_{c}} & \mathbf{0} \end{matrix}\right]^{-1}
\left(
\left[\begin{matrix}\frac{\partial\boldsymbol{\tau}}{\partial\mathbf{x}} \\ \frac{\partial\mathbf{a}_0}{\partial\mathbf{x}}\end{matrix}\right]\delta\mathbf{x} +
\left[\begin{matrix}\frac{\partial\boldsymbol{\tau}}{\partial\mathbf{u}} \\ \frac{\partial\mathbf{a}_0}{\partial\mathbf{u}}\end{matrix}\right]\delta\mathbf{u}
\right)\nonumber\\ &=&
\left[\begin{matrix}\mathbf{y_x} \\ -\mathbf{g_x}\end{matrix}\right]\delta\mathbf{x} +
\left[\begin{matrix}\mathbf{y_u} \\ -\mathbf{g_u}\end{matrix}\right]\delta\mathbf{u},
\end{eqnarray}
where $\frac{\partial\boldsymbol{\tau}}{\partial\mathbf{x}}$, $\frac{\partial\boldsymbol{\tau}}{\partial\mathbf{u}}$ are the \gls{rnea} derivatives, and $\frac{\partial\mathbf{a}_0}{\partial\mathbf{x}}$, $\frac{\partial\mathbf{a}_0}{\partial\mathbf{u}}$ are the kinematics derivatives of the frame acceleration~\cite{carpentier-rss18,carpentier-sii19}.
We use a LDU decomposition to invert the blockwise matrix\footnote{Note that this is the \gls{kkt} matrix.} in~\eref{eq:kkt_fwddyns_der}.

\subsubsection{Impulse dynamics}
We can similarly describe the impulse dynamics of a multibody system\footnote{Transitions from non-contact to contact condition~\cite{featherstone-rbdbook}.} as:
\begin{equation}
\left[\begin{matrix}\mathbf{M} & \mathbf{J}^{\top}_c \\ {\mathbf{J}_{c}} & \mathbf{0} \end{matrix}\right]
\left[\begin{matrix}\mathbf{v}^+ \\ -\boldsymbol{\Lambda} \end{matrix}\right] =
\left[\begin{matrix}\mathbf{M}\mathbf{v}^- \\ -e\mathbf{J}_c\mathbf{v}^- \\\end{matrix}\right],
\label{eq:kkt_impulse}
\end{equation}
where $e\in[0,1]$ is the restitution coefficient that considers compression / expansion, $\boldsymbol{\Lambda}$ is the contact impulse and, $\mathbf{v}^-$ and $\mathbf{v}^+$ are the discontinuous changes in the generalized velocity (i.e., velocity before and after impact, respectively).
Perfect inelastic collision produces a contact velocity equal to zero, i.e., $e=0$.
Similarly, we use the Cholesky decomposition to efficiently compute the impulse dynamics and its derivatives.

\section{Feasibility-prone Differential Dynamic Programming}\label{sec:fddp}
In this section, we describe our novel solver for multiple-shooting \gls{oc} called \acrfull{fddp}.
First, we briefly describe the \gls{ddp} algorithm~(\sref{sec:ddp}).
Then, we analyze the numerical behavior of classical multiple-shooting methods~(\sref{sec:multiple_shooting_gap}).
With this in mind, we propose a modification of the forward and the backward passes in~\sref{sec:fddp_rollout} and \ref{sec:fddp_backward}, respectively.
Finally, we propose a new model for the expected reduction cost and line-search procedure based on the Goldstein condition (\sref{sec:accepting_step}).

\subsection{Differential dynamic programming}\label{sec:ddp}
\gls{ddp} belongs to the family of \gls{oc} and indirect trajectory optimization methods~\cite{mayne-66}.
It locally approximates the optimal flow (i.e., the Value function) around $(\delta\mathbf{x}_k,\delta\mathbf{u}_k)$ as
\begin{equation}
 V_k(\delta\mathbf{x}_k) = \min\limits_{\delta\mathbf{u}_k} l_k(\delta\mathbf{x}_k,\delta\mathbf{u}_k) + V_{k+1}(\mathbf{f}_k(\delta\mathbf{x}_k,\delta\mathbf{u}_k)),
 \label{bellman}
\end{equation}
which breaks the \gls{oc} problem into a sequence of simpler subproblems by using  ``Bellman's principle of optimality'', i.e.:
\begin{eqnarray}\label{eq:Qeq}
	&&\hspace{-2em}\delta\mathbf{u}^*_k(\delta\mathbf{x}_k) = \\\nonumber
	&&\arg\min_{\delta\mathbf{u}_k} \overbrace{\frac{1}{2}
	\begin{bmatrix}
		1 \\ \delta\mathbf{x}_k \\ \delta\mathbf{u}_k
	\end{bmatrix}^T
	\begin{bmatrix}
		0 & \mathbf{Q}^T_{\mathbf{x}_k} & \mathbf{Q}^T_{\mathbf{u}_k} \\
		\mathbf{Q}_{\mathbf{x}_k} & \mathbf{Q}_{\mathbf{xx}_k} & \mathbf{Q}_{\mathbf{xu}_k} \\
		\mathbf{Q}_{\mathbf{u}_k} & \mathbf{Q}^T_{\mathbf{xu}_k} & \mathbf{Q}_{\mathbf{uu}_k}
	\end{bmatrix}
	\begin{bmatrix}
		1 \\ \delta\mathbf{x}_k \\ \delta\mathbf{u}_k
	\end{bmatrix}}^{\mathbf{H}(\delta\mathbf{x}_k,\delta\mathbf{u}_k,\bar{V}_k,k)}.
\end{eqnarray}
Note that $l_k(\cdot)$, $\mathbf{f}_k(\cdot)$ are the \gls{lq} approximation of the cost and dynamics functions, respectively; $\delta\mathbf{x}_k$, $\delta\mathbf{u}_k$ reflects the fact that we linearize the problem around a guess $(\mathbf{x}^i_k, \mathbf{u}^i_k)$.
This remark is particularly important (1) to understand our \gls{fddp} algorithm and (2) to deal with the geometric structure of dynamical systems\footnote{The configuration point lies on a manifold $Q$ (e.g., a Lie group) and the system derivatives lies in its tangent space.} (e.g. using symplectic integrators~\cite{hairer-bookgeom}).

The $\mathbf{Q}_{**}$ terms represent the \gls{lq} approximation of the control Hamiltonian function $\mathbf{H}(\cdot)$.
The solution of the entire \gls{oc} problem is computed through the Riccati recursion formed by sequentially solving~\eref{eq:Qeq}.
This procedure provides the feed-forward term $\mathbf{k}_k$ and feedback gains $\mathbf{K}_k$ at each discretization point $k$.

\subsection{The role of gaps in multiple-shooting}\label{sec:multiple_shooting_gap}
The multiple-shooting \gls{oc} formulation introduces intermediate states $\mathbf{x}_{k}$ (i.e., shooting nodes) as additional decision variables to the numerical optimization problem with extra equality constraints that attend to close the gaps\footnote{It is also called \textit{defects} in multiple-shooting literature.}~\cite{bock-ifac84}, i.e.
\begin{equation}\label{eq:gap}
	\mathbf{\bar{f}}_{k+1} = \mathbf{f}(\mathbf{x}_k,\mathbf{u}_k) - \mathbf{x}_{k+1},
\end{equation}
where $\mathbf{\bar{f}}_{k+1}$ represents the gap in the dynamics, $\mathbf{f}(\mathbf{x}_k,\mathbf{u}_k)$ is the rollout state at interval $k+1$, and $\mathbf{x}_{k+1}$ is the next shooting state (decision variable).
For the remainder of this paper, we assume that there is a shooting node for each integration step along the trajectory.

By approaching the direct multiple-shooting formulation as a \gls{sqp} problem, one can describe a single \gls{qp} iteration as
\begin{equation}
	\label{eq:multiple_shooting_oc_problem}
	\begin{aligned}
			\min_{\delta\mathbf{X},\delta\mathbf{U}}
			&
			& & l_N(\delta\mathbf{x}_k) +\sum_{k=0}^{N-1} l_k(\delta\mathbf{x}_k,\delta\mathbf{u}_k) \\
			& \hspace{-2em}\textrm{s.t.}
			& &   \hspace{-1em}\delta\mathbf{x}_0 = \tilde{\mathbf{x}}_0,\\
			& & & \hspace{-1em}\delta\mathbf{x}_{k+1} = \mathbf{f_x}_k\delta\mathbf{x}_k + \mathbf{f_u}_k\delta\mathbf{u}_k + \mathbf{\bar{f}}_{k+1},
	\end{aligned}
\end{equation}
where the \gls{sqp} sequentially builds and solves a single \gls{qp} problem until it reaches the convergence criteria.
The solution of \eref{eq:multiple_shooting_oc_problem} provides us a search direction.
Then, we can find a step length $\alpha$ for updating the next guess $(\mathbf{X}_{i+1},\mathbf{U}_{i+1})$ as
\begin{equation}\label{eq:sqp_iteration}
	\begin{bmatrix}
		\mathbf{X}_{i+1} \\ \mathbf{U}_{i+1}
	\end{bmatrix} =
	\begin{bmatrix}
		\mathbf{X}_{i} \\ \mathbf{U}_{i}
	\end{bmatrix} + \alpha
	\begin{bmatrix}
		\delta\mathbf{X}_{i} \\ \delta\mathbf{U}_{i}
	\end{bmatrix}
\end{equation}
where the new guess trajectory $\left(\mathbf{X}_{i+1},\mathbf{U}_{i+1}\right)$ does not necessarily close the gaps as we explain below.

\subsubsection{KKT problem of the multiple-shooting formulation}\label{sec:fddp_kkt}
To understand the behavior of the gaps, we formulate the \gls{kkt} problem in~\eref{eq:sqp_iteration} for a single shooting interval $k$ as:
\begin{eqnarray}\label{eq:multishooting_dualfeas}
	\overbrace{\begin{bmatrix}
		\mathbf{l}_{\mathbf{xx}_k} & \mathbf{l}_{\mathbf{xu}_k} \\
		\mathbf{l}^T_{\mathbf{xu}_k} & \mathbf{l}_{\mathbf{uu}_k}
	\end{bmatrix}}^{\mathbf{H}_k}
	\overbrace{\begin{bmatrix}
		\delta\mathbf{x}_k \\ \delta\mathbf{u}_k
	\end{bmatrix}}^{\delta\mathbf{w}_k} +
	\overbrace{\begin{bmatrix}
		\mathbf{I} & -\mathbf{f}^T_{\mathbf{x}_k} \\
					 & -\mathbf{f}^T_{\mathbf{u}_k}
	\end{bmatrix}}^{\left[\begin{smallmatrix}{\nabla\mathbf{g}^-_k} & {\nabla\mathbf{g}^+_k}\end{smallmatrix}\right]}
	\begin{bmatrix}
		\boldsymbol{\lambda}_k \\ \boldsymbol{\lambda}_{k+1}
	\end{bmatrix} = -
	\overbrace{\begin{bmatrix}
		\mathbf{l}_{\mathbf{x}_k} \\ \mathbf{l}_{\mathbf{u}_k}
	\end{bmatrix}}^{\nabla\boldsymbol{\Phi}_k},&
	\\\label{eq:multishooting_primalfeas}
	\begin{bmatrix}
		\mathbf{I} &  \\
		-\mathbf{f}_{\mathbf{x}_k} & -\mathbf{f}_{\mathbf{u}_k}
	\end{bmatrix}
	\begin{bmatrix}
		\delta\mathbf{x}_k \\
		\delta\mathbf{u}_k
	\end{bmatrix} =
	\begin{bmatrix}
		\mathbf{\bar{f}}_k \\ \mathbf{\bar{f}}_{k+1}
	\end{bmatrix},&
\end{eqnarray}
where \eref{eq:multishooting_dualfeas}, (\ref{eq:multishooting_primalfeas}) are the dual and primal feasibility of the \gls{fonc} of optimality, respectively.
The Jacobians and Hessians of the cost function (\gls{lq} approximation) are $\mathbf{l_x}$, $\mathbf{l_u}$, and $\mathbf{l_{xx}}$, $\mathbf{l_{xu}}$, $\mathbf{l_{uu}}$, respectively.
The Lagrangian multipliers of the \gls{kkt} problem are $(\boldsymbol{\lambda}_k,\boldsymbol{\lambda}_{k+1})$.

We obtain the search direction $\delta\mathbf{w}_k$ by solving the \gls{fonc} as follows:
\begin{equation}\label{eq:gap_prediction}
	\begin{bmatrix}
		\delta\mathbf{w}_k \\ \delta\boldsymbol{\lambda}_k \\ \delta\boldsymbol{\lambda}_{k+1}
	\end{bmatrix} =
	\begin{bmatrix}
		\mathbf{H}_k & {\nabla\mathbf{g}^-_k} & {\nabla\mathbf{g}^+_k} \\
		\nabla\mathbf{g}^{-^T}_k & \\
		\nabla\mathbf{g}^{+^T}_k &
	\end{bmatrix}^{-1}
	\begin{bmatrix}
		\nabla\boldsymbol{\Phi}_k \\ \mathbf{\bar{f}}_k \\ \mathbf{\bar{f}}_{k+1}
	\end{bmatrix},
\end{equation}
in which we note that a $\alpha$-step closes the gap at $k$ by a factor of $(1-\alpha)\mathbf{\bar{f}}_k$, while only a full-step $(\alpha=1)$ can close the gap completely.
Below, we explain how to ensure this multiple-shooting behavior in the forward-pass.

\subsection{Nonlinear rollout avoids merit function}\label{sec:fddp_rollout}
\gls{sqp} often requires a \emph{merit} function to compensate the errors that arise from the local approximation of the classical line-search.
Defining a suitable merit function is often challenging, which is why we do not follow this approach.
Instead, we avoid (a) the linear-prediction error of the dynamics -- i.e. search direction defined by \eref{eq:gap_prediction} -- with a nonlinear rollout and (b) the requirement of a merit function.

For a nonlinear rollout, the prediction of the gaps after applying an $\alpha$-step is:
\begin{eqnarray}\label{eq:nonlinear_gap_pred}
	\mathbf{\bar{f}}_{k+1}^{i+1} &=& \mathbf{\bar{f}}_{k+1}^{i} - \alpha(\delta\mathbf{x}_{k+1} - \mathbf{f_x}_k\delta\mathbf{x}_k - \mathbf{f_u}_k\delta\mathbf{u}_k)\nonumber\\
	&=& (1-\alpha)(\mathbf{f}(\mathbf{x}_k,\mathbf{u}_k)-\mathbf{x}_{k+1}),
\end{eqnarray}
and we maintain the same gap contraction rate of the search direction \eref{eq:gap_prediction}.
Therefore, we have the following rollout:
\begin{eqnarray}
	\mathbf{\hat{x}}_0 &=& \mathbf{\tilde{x}}_0 - (1 - \alpha)\mathbf{\bar{f}}_0, \nonumber\\
	\mathbf{\hat{u}}_k &=& \mathbf{u}_k + \alpha\mathbf{k}_k + \mathbf{K}_k(\mathbf{\hat{x}}_k-\mathbf{x}_k), \\\nonumber
	\mathbf{\hat{x}}_{k+1} &=& \mathbf{f}_k(\mathbf{\hat{x}}_k,\mathbf{\hat{u}}_k) - (1 - \alpha)\mathbf{\bar{f}}_k,
\end{eqnarray}
where $\mathbf{k}_k$ and $\mathbf{K}_k$ are the feed-forward term and feedback gains computed during the backward pass, respectively.
Note that the forward pass of the classical \gls{ddp} always closes the gaps, and with $\alpha=1$, the \gls{fddp} forward pass behaves exactly as the classical \gls{ddp} one.

\subsection{Backward pass under an infeasible guess trajectory}\label{sec:fddp_backward}
Gaps in the dynamics and infeasible warm-starts generate derivatives at different points.
The Riccati recursion updates the Value and Hamiltonian functions based on these derivatives.
The classical \gls{ddp} algorithm overcomes this problem by first performing an initial forward pass.
However, from a theoretical point, it corresponds to only being able to warm-start the solver with the control trajectory $\mathbf{U}_0$, which is not convenient in practice\footnote{It is straight-forward to obtain a state trajectory $\mathbf{X}_0$ that provides an initial guess for the \gls{oc} solver, however, establishing a corresponding control trajectory $\mathbf{U}_0$ beyond quasi-static maneuvers is a limiting factor.}.

We adapt the backward pass to accept infeasible guesses as proposed by~\cite{giftthaler-iros18}.
It assumes a \gls{lq} approximation of the Value function, i.e. the Hessian is constant and the Jacobian varies linearly.
We use this fact to map the Jacobians and Hessian of the Value function from the next shooting-node to the current one.
Therefore, the Riccati recursions are modified as follows:
\begin{eqnarray}\label{lastQeq}\nonumber
	\mathbf{Q}_{\mathbf{x}_k} & = & \mathbf{l}_{\mathbf{x}_k} + \mathbf{f}^T_{\mathbf{x}_k} V^+_{\mathbf{x}_{k+1}}, \\\nonumber
	\mathbf{Q}_{\mathbf{u}_k} & = & \mathbf{l}_{\mathbf{u}_k} + \mathbf{f}^T_{\mathbf{u}_k} V^+_{\mathbf{x}_{k+1}}, \\
	\mathbf{Q}_{\mathbf{xx}_k} & = & \mathbf{l}_{\mathbf{xx}_k} + 
	\mathbf{f}^T_{\mathbf{x}_k} V_{\mathbf{xx}_{k+1}} \mathbf{f}_{\mathbf{x}_k},\\\nonumber
	\mathbf{Q}_{\mathbf{xu}_k} & = & \mathbf{l}_{\mathbf{xu}_k} + 
	\mathbf{f}^T_{\mathbf{x}_k} V_{\mathbf{xx}_{k+1}} \mathbf{f}_{\mathbf{u}_k},\\\nonumber
	\mathbf{Q}_{\mathbf{uu}_k} & = & \mathbf{l}_{\mathbf{uu}_k} + 
	\mathbf{f}^T_{\mathbf{u}_k} V_{\mathbf{xx}_{k+1}} \mathbf{f}_{\mathbf{u}_k}.
\end{eqnarray}
where $V^+_{\mathbf{x}_{k+1}}=V_{\mathbf{x}_{k+1}}+V_{\mathbf{xx}_{k+1}}\mathbf{\bar{f}}_{k+1}$ is the Jacobian of the Value function after the deflection produced by the gap $\mathbf{\bar{f}}_{k+1}$, and the Hessian of the Value function remains unchanged.

\subsection{Accepting a step}\label{sec:accepting_step}
The expectation of the total cost reduction proposed by~\cite{tassa-iros12} does not consider the deflection introduced by the gaps.
This is a critical point to evaluate the success of a trial step during the numerical optimization.
From our line-search procedure, we know that the expected reduction on the cost has the form:
\begin{equation}
	\Delta J(\alpha) = \Delta_1\alpha + \frac{1}{2}\Delta_2\alpha^2,
\end{equation}
where, by closing the gaps as predicted in~\eref{eq:gap_prediction} in the linear rollout, we obtain:
\begin{eqnarray}
	\Delta_1 = \sum_{k=0}^{N-1} \mathbf{k}_k^\top\mathbf{Q}_{\mathbf{u}_k} +\mathbf{\bar{f}}_k^\top(V_{\mathbf{x}_k} - V_{\mathbf{xx}_k}\mathbf{x}_k),\nonumber\\
	\Delta_2 = \sum_{k=0}^{N-1} \mathbf{k}_k^\top\mathbf{Q}_{\mathbf{uu}_k}\mathbf{k}_k + \mathbf{\bar{f}}_k^\top(2 V_{\mathbf{xx}_k}\mathbf{x}_k - V_{\mathbf{xx}_k}\mathbf{\bar{f}}_k).
\end{eqnarray}
Note that if all gaps are closed, then this expectation model matches the one reported in~\cite{tassa-iros12}.

We use the Goldstein condition to check for the trial step, instead of the Armijo condition typically used in classical \gls{ddp} algorithms, e.g.,~\cite{tassa-iros12}.
The reason is due to the fact that $\Delta J$ might be an ascent direction, for instance, during the infeasible iterations.
Therefore, \gls{fddp} accepts the step if the cost reduction is:
\begin{equation}
	 l'-l \le
	\begin{cases}
  b_1 \Delta J(\alpha) & \textrm{if }\Delta J(\alpha)\le 0 \\
  b_2 \Delta J(\alpha) & \textrm{otherwise}
	\end{cases}
\end{equation}
where $b_1$, $b_2$ are adjustable parameters, we used in this paper $b_1 = 0.1$ and $b_2 = 2$.
This critical mathematical aspect has not been considered in~\cite{giftthaler-iros18}.

\section{Results}
In this section, we show the capabilities of our multi-contact optimal control framework.
We first compute various legged gaits for both quadruped and biped robots (\sref{sec:gaits}).
As our formulation is simple and does not depend on a good initial guess, it can be used easily with different legged robots.
Next, we analyze the performance of the \gls{fddp} with the generation of highly-dynamic maneuvers such as jumps and front-flips.
These motions are computed within a few iterations and milliseconds as reported. 
We have deliberately ignored friction-cone constraints and torque limits for the sake of evaluating the \gls{fddp}, however, it is possible to include those inequality constraints through quadratic penalization as shown in the cover clip of accompanying video.
The accompanying video\footnote{\url{https://youtu.be/wHy8YAHwj-M}.} highlights all different motions reported in this section.

\subsection{Various legged gaits}\label{sec:gaits}
We computed different gaits --- walking, trotting, pacing, and bounding --- with our \gls{fddp} algorithm in the order of milliseconds. 
All these gaits are a direct outcome of our algorithm given a predefined sequence of contacts and step timings.
These motions are computed in around 12 iterations. 
We used the same weight values and cost functions for all the quadrupedal gaits, and similar weight values for the bipedal walking.

The cost function is composed of the \gls{com} and the foot placement tracking costs together with regularization terms for the state and control.
We used piecewise-linear functions to describe the reference trajectory for the swing foot.
Additionally, we strongly penalize footstep deviation from the reference placement.
We warm-start our solver using a linear interpolation between the nominal body postures of a sequence of contact configurations.
This provides us a set of body postures together with the nominal joint postures as state warm-start $\mathbf{X}_0$.
Then, the control warm-start $\mathbf{U}_0$ is obtained by applying the quasi-static assumption\footnote{The quasi-static torques are numerically computed through Newton steps using the reference posture as an equilibrium point.} along $\mathbf{X}_0$.


In each switching phase\footnote{In this work, with ``switching phases'' we refer to contact gain.}, we use the impulse dynamics to ensure the contact velocity equals zero, see \eref{eq:kkt_impulse}.
We observed that the use of impulse models improves the algorithm convergence compared to penalizing the contact velocity.
We used a weighted least-squares function to regularize the state with respect to the nominal robot posture, and quadratic functions for the tracking costs and control regularization.


\begin{figure}[htb]
    \centering
    \includegraphics[width=0.85\linewidth]{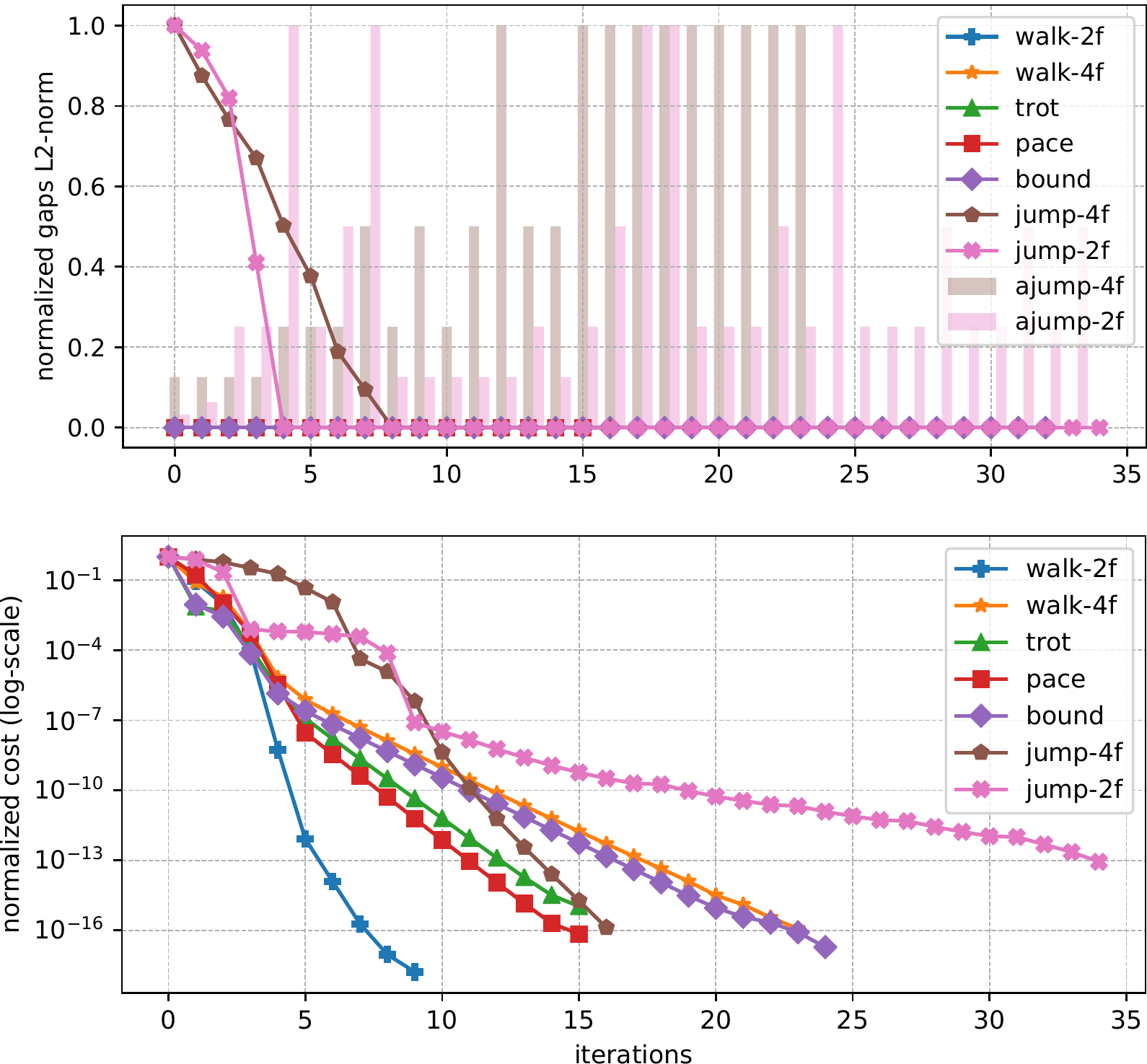}\vspace{-0.8em}
    \caption{Gaps contraction, step length, and convergence rates for different motions.
    (top) Gaps are closed in the first iteration for simpler motions such as biped walking and quadrupedal gaits.
    Instead, the \gls{fddp} solver chooses to keep the gaps open for the early iterations for highly-dynamic maneuvers.
    Note that we use the L2-norm of the total gaps, i.e., gaps for all the nodes of the trajectory.
    (bottom) The required iterations increases mainly with the dynamics of the gait and numbers of nodes.
    For instance, we can see lower rate of improvement in the first nine iterations in the ANYmal (jump-4f) and ICub jumps (jump-2f).
    In case of the quadrupedal walking, we have very short durations in the four-feet support phases, making it a \emph{dynamic} walk.}
    \label{fig:convergence_rate}\vspace{-1em}
\end{figure}

\begin{figure*}[htb]
    \centering
    \href{https://youtu.be/wHy8YAHwj-M?t=74}{\includegraphics[width=0.93\textwidth]{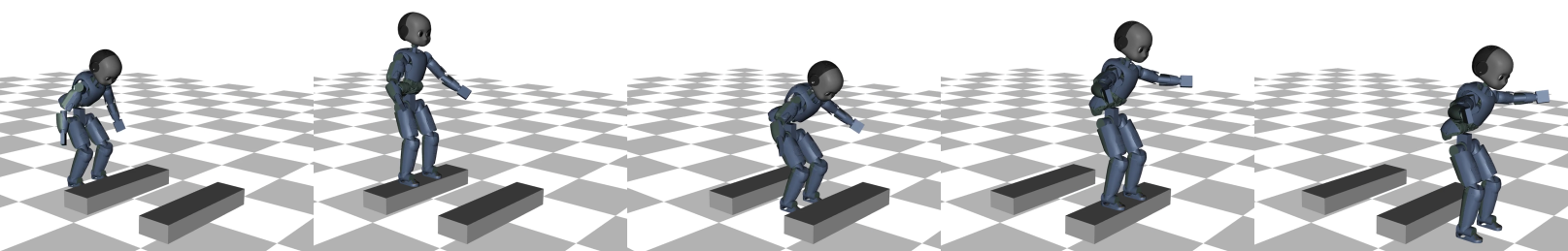}}
    \href{https://youtu.be/wHy8YAHwj-M?t=26}{\includegraphics[width=0.93\textwidth]{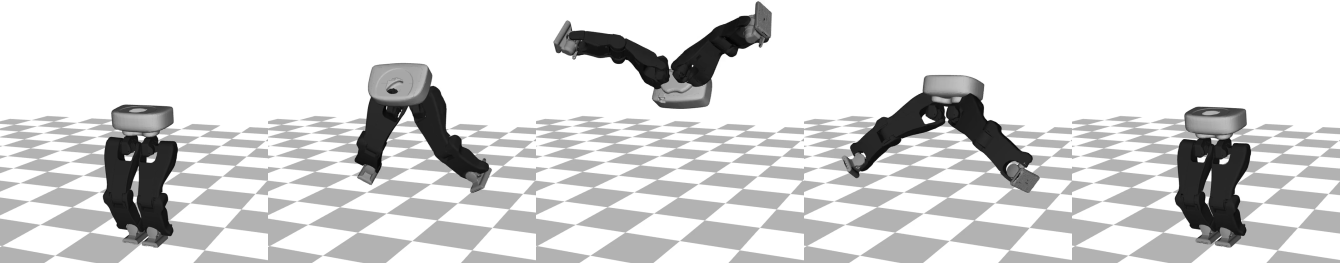}}
    \href{https://youtu.be/wHy8YAHwj-M?t=46}{\includegraphics[width=0.93\textwidth]{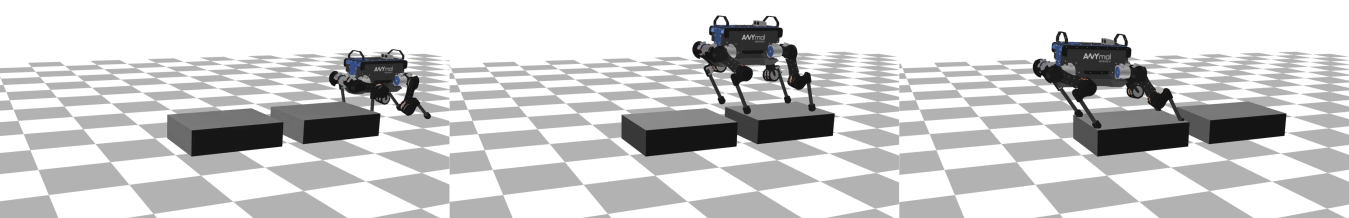}}
    \caption{Snapshots of generated highly-dynamic maneuvers in legged robots using the feasibility-prone differential dynamic algorithm.
    (top) jumping obstacles in a humanoid robot; (middle) front-flip maneuver in a biped robot; (bottom) jumping obstacles in a quadruped robot.}
    \label{fig:highly_dyn_maneuvers}
\end{figure*}

\subsection{Highly-dynamic maneuvers}\label{sec:maneuvers}
Our \gls{fddp} algorithm is able to compute highly-dynamic maneuvers such as front-flip and jumping in the order of milliseconds (\fref{fig:highly_dyn_maneuvers}).
These motions are often computed in between 12--36 iterations with a na\"ive and infeasible $\mathbf{X}_0,\mathbf{U}_0$ warm-start.
We used the same initialization, weight values and cost functions reported in~\sref{sec:gaits}, with a slightly incremented weight for the state regularization during the impact phases (i.e. $w_{xReg}=10$).
Additionally, and for simplicity, we included a cost that penalizes the body orientation in the ICub jumps.
Similarly to other cost functions, we used a quadratic penalization with a weight value of $10^4$.
Note that a more elaborate cost function could be incorporated: arm motions, angular momentum regulation, etc.

The advantage of our \gls{fddp} algorithm is clearly evident in the generation of highly-dynamic maneuvers, where feasible rollouts might produce trajectories that are unstable and far from the solution.
The classical \gls{ddp} has a poor globalization strategy that comes from inappropriate feasible rollouts in the first iterations; it struggles to solve these kind of problems.


\subsection{Runtime, contraction, and convergence}
We analyzed the gaps contraction and convergence rates for all the presented motions. 
To easily compare the results, we normalize the gaps and cost values per each iteration as shown in~\fref{fig:convergence_rate}.
We use the L2-norm of the total gaps and plot the applied step-length for the jumping motions (ajump-4f and ajump-2f).
These results show that keeping the gaps open is particularly important for highly-dynamic maneuvers such as jumping.
Indeed, in the jumping motions, \gls{fddp} keeps the gaps open for few iterations.
Additionally, we often observed in practice super-linear convergence of the \gls{fddp} algorithm after closing the gaps.
This is expected since the \gls{fddp} forward-pass behaves as the \gls{ddp} forward pass when the gaps are closed, which is defined by the search direction of a multiple-shooting formulation with only equality constraints (\sref{sec:fddp_kkt}).

Highly-dynamic maneuvers have a lower rate of improvement in the first iterations, cf.~\fref{fig:convergence_rate}~(bottom).
The same occurs in the quadrupedal walking case (walk-4f), in which the four-feet support phases have a very short duration ($\Delta t = 2$ \si{\milli\second}).
Our \gls{fddp} algorithm, together with the impact models, shows competitive convergence rates when compared to the reported results in~\cite{neunert-ral17,farshidian-icra17}, respectively.

The motions converge within 10 to 34 iterations, with an overall computation time of less than \SI{0.5}{\second}.
The numerical integration step size is often $\delta t = 1\times10^{-2}$~\si{\second}, with the exception of the biped walking $\delta t = 3\times 10^{-2}$~\si{\second}, and the number of nodes are typically between 60 to 115.
Therefore, the optimized trajectories have a horizon of between \SIrange{0.6}{3}{\second}.

\begin{figure}[htb]
    \centering
    \includegraphics[width=0.8\linewidth,trim={0.3cm 0.3cm 0.3cm 0.3cm},clip]{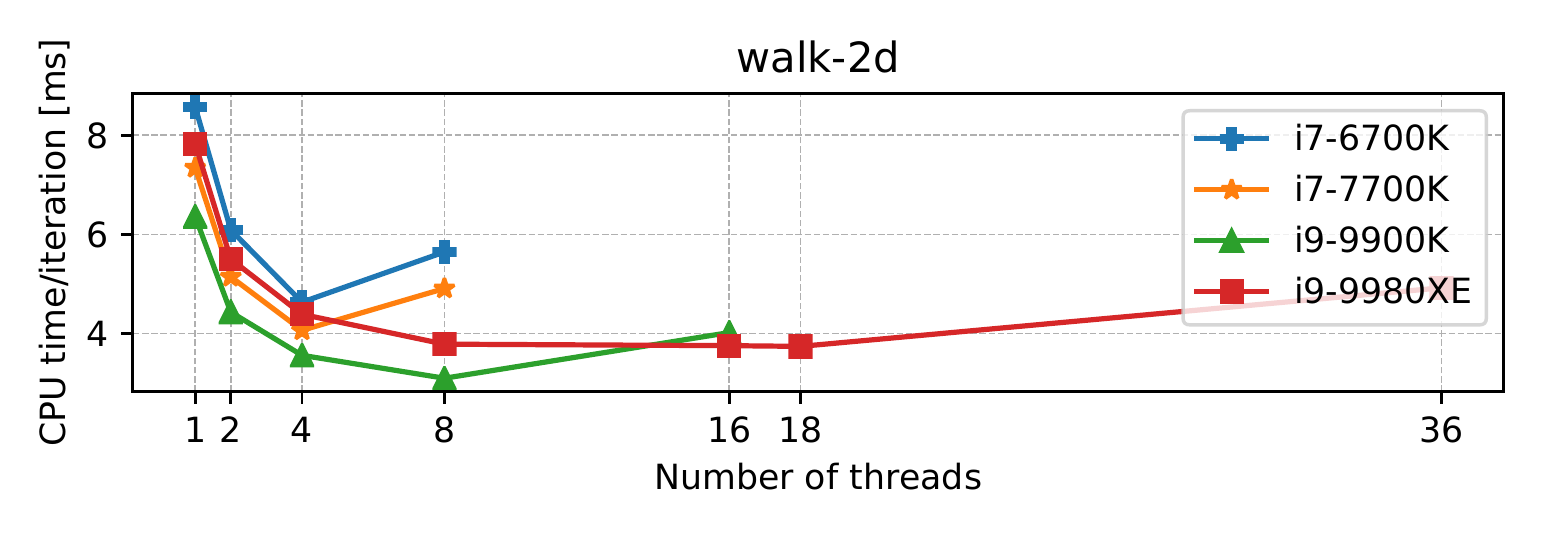}\vspace{-0.8em}
    \caption{Computation time per iteration for different CPUs and level of parallelism. Note that the use of hyper-threading decreases the computation frequency for all tested CPUs.}
    \label{fig:optimal_num_threads}
\end{figure}

\begin{figure}[htb]
    \centering
    \includegraphics[width=0.85\linewidth,trim={0.3cm 0.3cm 0.3cm 0.3cm},clip]{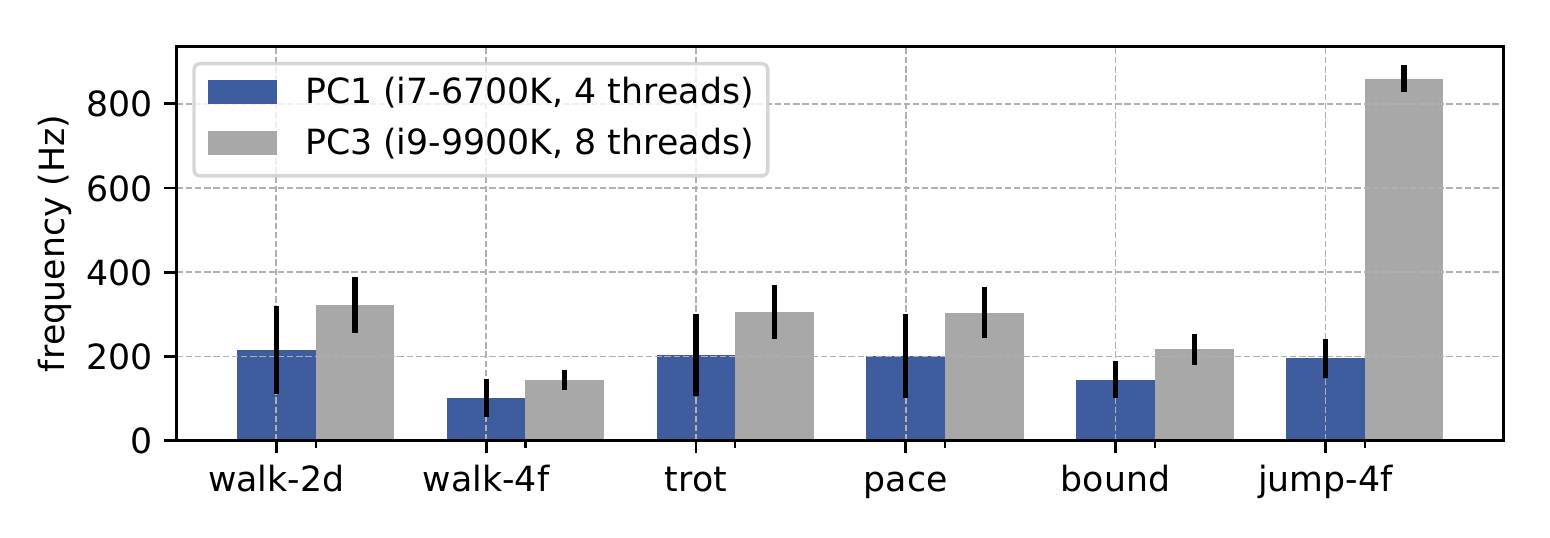}\vspace{-0.8em}
    \caption{Computation frequency per iteration for different motions for different PCs.
    PC1 has specifications typical for on-board computers found on robots, while PC3 uses high-performance CPU and RAM.
    The reported values use the optimal number of threads as identified in \fref{fig:optimal_num_threads}.}
    \label{fig:computation_time}\vspace{-1em}
\end{figure}

We also benchmark the computation time for a single iteration using our solver.
The number of contacts does not affect the computation time; it scales linearly with respect to the number of nodes.
With multi-threading, our efficient implementation of contact dynamics achieves computation rates up to \SI{859.6}{\hertz} (jump-4f on i9-9900K, 60 nodes).
We parallelize only the computation of the derivatives, and roughly speaking, we reduce the computation time in half using four to eight threads (cf. \fref{fig:optimal_num_threads}).
To understand the performance of \acrshort{crocoddyl}, we have run 50000 trials, for each of the benchmark motions, on four different Intel PCs with varying levels of parallelization\footnote{\emph{PC1}: i7-6700K~@~$4.00\si{\giga\hertz}\times 8$ with 32~$\si{\giga\byte}$ $2133\si{\mega\hertz}$ RAM, \emph{PC2}: i7-7700K~@~$4.20\si{\giga\hertz}\times 8$ with 16~$\si{\giga\byte}$ $2666\si{\mega\hertz}$ RAM, \emph{PC3}: i9-9900K~@~$3.60\si{\giga\hertz}\times 16$ with 64~$\si{\giga\byte}$ $3000\si{\mega\hertz}$ RAM, and \emph{PC4}: i7-9900XE~@~$3.00\si{\giga\hertz}\times 36$ with 128~$\si{\giga\byte}$ $2666\si{\mega\hertz}$ RAM.}.
We used the optimal number of threads for each PC as identified in \fref{fig:optimal_num_threads}.
The computation frequency per one iteration is reported in~\fref{fig:computation_time}.

\section{Conclusion}
We presented a novel and efficient framework for multi-contact optimal control.
The gap contraction of \gls{fddp} is equivalent to direct multiple-shooting formulations with only equality constraints (i.e. the Newton method applied to the \gls{kkt} conditions).
However, and in contrast to classical multiple-shooting, \gls{fddp} does not add extra decision variables which often increases the computation time per iteration due to factorization; it has cubic complexity in matrix dimension.
\gls{fddp} also improves the poor globalization strategy of classical \gls{ddp} methods.
This allows us to solve highly-dynamic maneuvers such as jumping and front-flip in the order of milliseconds.
Thanks to our efficient method for computing the contact dynamics and their derivatives, we can solve the optimal control problem at high frequencies.
Finally, we demonstrated the benefits of using impact models for contact gain phases.
Our core idea about feasibility could incorporate inequality constraints in the form of penalization terms.
Future work will focus on feasibility under inequality constraints such as torque limits, and friction cone.
Those inequalities constraints can be handled using interior-point~\cite{xie-icra17} or Augmented Lagrangian~\cite{howell-19} methods.

\addtolength{\textheight}{-7.cm}   

\bibliographystyle{./IEEEtran}
\bibliography{IEEEabrv,IEEEconf,references}

\end{document}